# Les représentations génétiques d'objets : simples analogies ou modèles pertinents ?

## *Le point de vue de l' « évolutique »*

**Laurent Krähenbühl**

*Centre de Génie Electrique de Lyon – UMR CNRS 5005*
*Ecole Centrale de Lyon – 36 avenue Guy de Collongue – 69134 Ecully CEDEX*
*Laurent.Krahenbuhl@ec-lyon.fr*

RÉSUMÉ. *Depuis une trentaine d'années, des analogies avec l'évolution naturelle (darwinisme) sont couramment utilisées pour optimiser des dispositifs techniques. Le plus souvent, ces méthodes « génétiques » ou « évolutionnaires » sont considérées uniquement du point de vue pratique, comme des méthodes d'optimisation performantes, qu'on peut utiliser à la place d'autres méthodes (gradients, simplexes, …), ou en association avec elles. A la suite de quelques auteurs célèbres (dont John H. Holland lui-même) nous essayons de montrer que les sciences et les techniques (mais aussi les organisations humaines, et plus généralement tous les systèmes complexes), obéissent effectivement à des lois d'évolution dont la génétique est un bon modèle représentatif, même si gênes et chromosomes sont « virtuels » : ainsi, loin d'être seulement un outil ponctuel d'aide à la synthèse de solutions technologiques, la représentation génétique est-elle un modèle dynamique global de l'évolution du monde façonné par l'agitation humaine. Ce point de vue sur l'évolution du monde technique et scientifique, que nous pourrions appeler « évolutique », peut simplement amuser ; il peut aussi interroger, et conduire (peut-être) à quelques conséquences pratiques.*

ABSTRACT. *For thirty years, analogy with natural evolution is commonly used to optimize technical devices. More often that not, these "genetic" or "evolutionary" methods are only view as practical and efficient tools, which could replace other optimization techniques (gradient methods, simplex, …) or could be used in association with them. In this paper, we try to show that sciences, techniques, human organizations, and more generally all complex systems, obey to evolution rules, whose the genetic is a good representative model, even if genes and chromosomes are " virtual". Thus, the genetic representation is not only a specific tool helping for the design of technological solutions, but also a global and dynamic model for the action of the human agitation on our world. This point of view on the evolution of sciences and techniques, which might be called "evolutic", can amuse or surprise; it could also open some ways for future researches.*

MOTS-CLÉS : *optimisation, heuristique, évolutique, évolutionnaire, génétique, évolution, coévolution, technique, scientifique, systèmes complexes, dynamique, théorie des schémas.*

**Préliminaire.**

En 1975, John Holland publie un livre magistral, « Adaptation in natural and artificial systems » (Holland, 1975), basé sur des travaux qu'il mène depuis 1962 et passé dans un premier temps totalement inaperçu. Son élève D. Goldberg publie à son tour en 1989 un ouvrage (Goldberg, 1989) sur les algorithmes génétiques appliqués à des problèmes concrets d'optimisation : ce qu'il propose (avec listage des codes !) « marche » si bien que ces 2 livres, leurs auteurs et cette famille de méthodes d'optimisation sont vite célèbres. L'œuvre de Holland se trouve ainsi en partie popularisée par rebond[1] : on associe son nom à ces méthodes magiques qui permettent de trouver le minimum d'une fonction à paramètres multiples.

Pourtant, la vision qu'il proposait de l'adaptation des systèmes artificiels, et de ses analogies avec celle des systèmes naturels, ne se limitait pas à cette problématique étroite ; cela est nettement souligné en 1995, dans une réédition augmentée. Holland y évoque l'intérêt des méthodes d'inspiration génétique pour l'étude *des systèmes adaptatifs complexes*, systèmes dont :   « [le] rôle est crucial pour une importante gamme d'activités humaines ». Cette conviction est aussi la nôtre, et notre ambition au travers de cet article est de tenter de la faire partager à nos lecteurs.

Afin que notre thèse soit abordable aussi pour les lecteurs peu familiarisés avec les méthodes d'optimisation, nous commencerons par présenter comment nous comprenons les méthodes génétiques après une quinzaine d'années de pratique, essentiellement dans le domaine du génie électrique.

La métaphore génétique, au sens du Darwinisme, est évoquée par tous les auteurs, pour expliquer le fonctionnement de ces méthodes[2]. Dans une seconde partie, nous montrerons que, si elles empruntent leur vocabulaire à la biologie, ces méthodes simulent en réalité le véritable processus d'évolution  (ou coévolution) technologique, ou scientifique, économique, social, tel qu'il se déroule effectivement sous nos yeux, en vraie grandeur. Ce sera l'occasion de préciser quelques propriétés d'une fonction coût, et le rôle que peut jouer la dynamique de l'évolution.

Nous terminerons en essayant de rétablir le lien avec le problème canonique d'optimisation, puis en proposant quelques pistes pour une recherche interdisciplinaire sur l'évolution et l'optimisation des systèmes complexes.

Au passage, nous reviendront ici ou là à certaines des idées de Holland (en particulier à la théorie des schémas, ou à des concepts qu'on retrouve dans d'autres domaines des sciences, comme sa notion de *building blocks*).

---

[1] si ce premier livre de Holland est beaucoup cité, le plus souvent en même temps que celui de Goldberg, il n'est pas certain qu'il ait été autant lu !

[2] d'autres méthodes leur sont contemporaines (algorithmes évolutionnaires) ou sont nées depuis, qui utilisent d'autres métaphores « naturelles » (essaims particulaires, fourmis, …)

# 1 Approche génétique du problème « canonique » d'optimisation.

## 1.1 La simulation dans la chaîne de conception.

L'ingénieur n'a pas seulement besoin de concevoir (c'est-à-dire proposer un dispositif remplissant des fonctions données), il doit en plus le faire de manière optimale, c'est-à-dire en minimisant un « coût » (terme générique pour un objectif à minimiser), ou en maximisant une performance, et en respectant un certain nombre de contraintes (par exemple, les dimensions d'un transformateur de centrale nucléaire doivent être compatibles avec le transport ferroviaire). Le plus souvent, l'ingénieur procède dans une première étape par essais-erreurs pour définir les grandes lignes d'une solution (technologie utilisée, structure, …) à partir de l'expérience accumulée et de modèles simples ; il se retrouve dans une seconde étape devant un dispositif défini par un jeu de paramètres (dimensions géométriques, caractéristiques de matériaux, valeurs d'excitations, paramètres de régulation, choix discret et limité de constituants dans un catalogue, …) qu'il doit choisir le mieux possible. En admettant qu'il dispose du modèle analytique ou numérique adéquat, il peut alors simuler le fonctionnement de son dispositif pour chaque choix particulier de ces paramètres, et déterminer comment cette réalisation virtuelle du dispositif va se comporter par rapport à l'objectif qu'il est chargé d'optimiser (Figure 1).

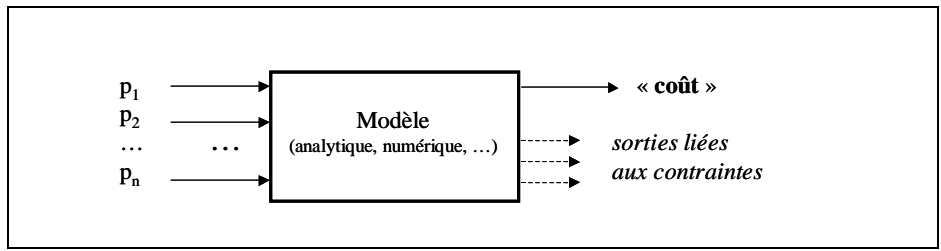

**Figure 1.** *Représentation du dispositif à optimiser sous forme de « boite noire »*

## 1.2 Optimiser : les méthodes locales déterministes.

Ce problème « direct » est généralement assez bien maîtrisé, mais ce n'est pas celui qui l'intéresse in fine.

Il est parfois possible d'inverser explicitement le modèle, ce cas heureux et en pratique très rare ne nous intéresse pas ici. De même, lorsqu'il n'y a qu'un ou deux paramètres en entrée, il est possible de « tracer » de manière exhaustive la réponse en fonction de ces entrées, et de repérer ainsi les valeurs optimales de manière quasi exacte (à la main, ou avec des méthodes itératives simples et classiques). Lorsque le nombre de paramètres est plus important, il n'est plus envisageable d'explorer l'espace de recherche par un quadrillage systématique (10 paramètres avec 10 valeurs chacun donneraient $10^{10}$ évaluations de la fonction coût …).

Toutes les techniques utilisées pour ces cas difficiles font alors appel d'une manière ou d'une autre à des essais successifs (Figure 2). Les différentes méthodes proposées se distinguent les unes des autres par les stratégies mises en œuvre pour que ce processus profite le plus possible de l'expérience accumulée, c'est-à-dire du résultat des essais précédents, de manière à accélérer la convergence vers une solution acceptable. D'une manière très générale, il s'agit donc, à partir des résultats déjà connus (c'est-à-dire d'un échantillonnage très partiel), d'obtenir une approximation vraisemblable de la fonction coût.

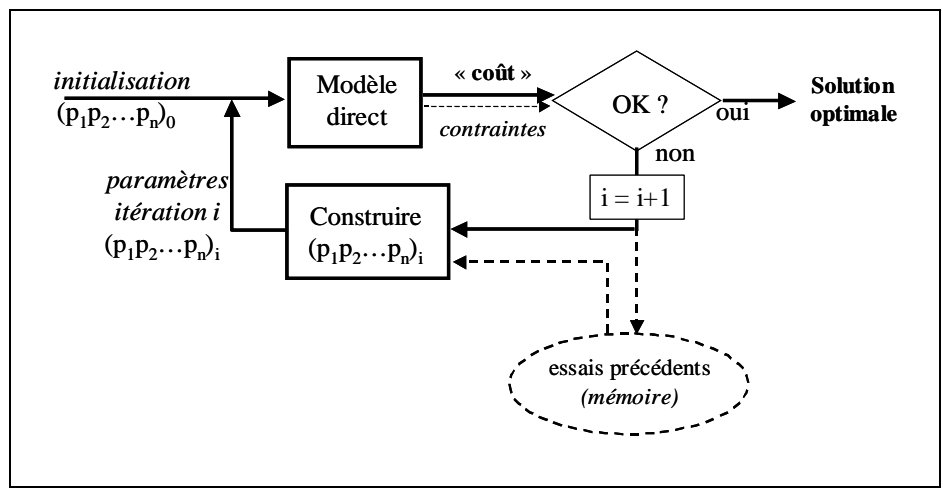

**Figure 2 :** *Méthode d'optimisation par essais-erreurs d'une boite noire paramétrée*

Comme il est représenté ici, ce processus itératif est séquentiel : à chaque itération correspond une seule évaluation de la fonction coût, qui est « parcourue » pas à pas suivant un certain cheminement, les différentes stratégies possibles (à définir dans la boîte « construire ») tendant toutes à chercher la direction descendante[3]. On utilise dans ce but quelques valeurs de l'historique (les dernières) dont on essaie d'extraire une information sur le gradient ou sur la courbure locale de la fonction coût. Ces méthodes itératives locales sont généralement complètement déterministes ; cela ne signifie pas que l'algorithme va systématiquement conduire au résultat exact, mais seulement que deux exécutions successives avec le même point d'origine donneront le même résultat.

---

[3] descendante pour la minimisation d'une fonction coût, ascendante pour la maximisation d'une performance. Les deux cas sont évidemment équivalents, à un renversement près de la fonction objectif.

Si la fonction coût a de bonnes propriétés (portant sur des critères comme sa continuité, sa dérivabilité, sa convexité, …), il est possible de construire suivant ce schéma des algorithmes dont la convergence vers un minimum (qui peut n'être que local, voir Figure 3, à droite) est démontrable.

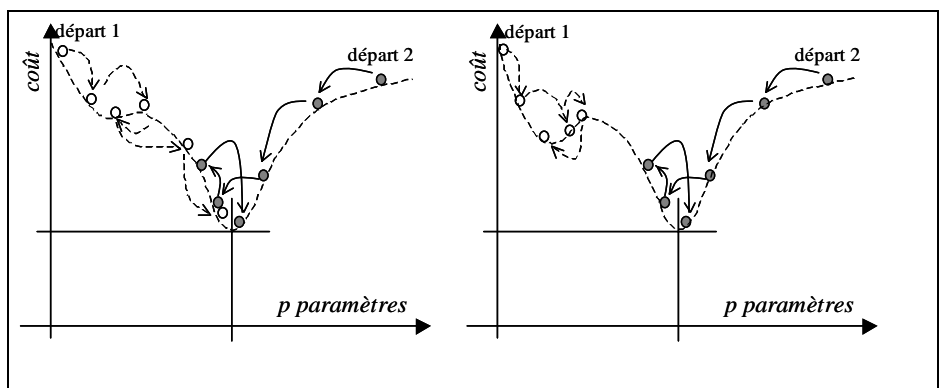

**Figure 3 :** *Progressions séquentielles vers un optimum, pour 2 points initiaux différents et deux fonctions coût différentes.*

### 1.3 Les méthodes stochastiques à population : généralités.

La classe de méthodes, dites « génétiques », sur laquelle nous portons ici notre attention, conserve le même principe général d'essais-erreurs, avec quelques aménagements :

- elle travaille en parallèle sur un ensemble de N jeux de paramètres candidats (typiquement, N=50), ensemble appelé « population ». A chaque itération de la méthode, on évalue donc N « individus », et c'est une nouvelle « génération », formée de N propositions nouvelles, qui est transmise pour l'itération suivante (Figure 4).

- la méthode fait à toutes les étapes appel au hasard. En particulier, la population initiale est tirée aléatoirement, avec une densité de probabilité uniforme sur tout l'espace de recherche (cet espace est défini par les variations acceptables des p paramètres de l'objet étudié).

- l'historique de la recherche complète n'est pas stocké : la mémoire des générations précédentes doit être traduite uniquement dans le choix judicieux de la génération suivante (qui contiendra donc une partie de la population précédente, plus des individus nouveaux).

- le choix des individus nouveaux de la génération suivante est lui aussi aléatoire. Cependant, l'algorithme mis en place pour les choisir doit, ou bien simplement augmenter la densité de peuplement vers ceux des individus existants qui ont les meilleures performances, ou même chercher à deviner où pourrait se « cacher » une zone encore plus favorable : la population va ainsi au fil des itérations se centrer d'elle-même autour du meilleur optimum connu, tout en continuant à prospecter dans des zones « prometteuses » (Figure 5). On ajoute même généralement une petite dose de recherche purement aléatoire.

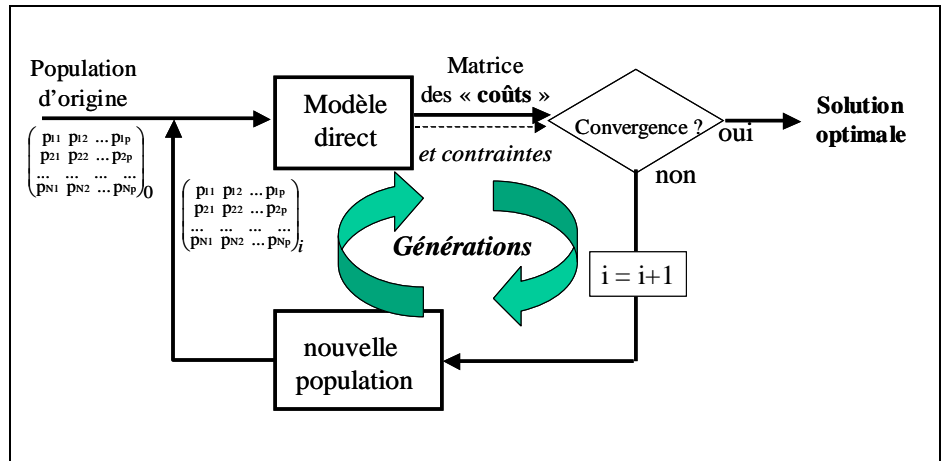

**Figure 4 :** *Méthode d'optimisation d'une boîte noire paramétrée, à l'aide de générations successives d'une population de N jeux de p paramètres.*

On imagine la difficulté psychologique que peut rencontrer un ingénieur à l'idée de faire appel à un processus largement aléatoire pour résoudre un problème défini par un cahier des charges pourtant précis et cartésien. Généralement, cette difficulté disparaît dès les premiers essais qu'il peut faire lui-même des méthodes génétiques, par exemple sur le problème combinatoire difficile du voyageur de commerce (Rennard, 2002)[4], car si l'ingénieur aime la rigueur, il est aussi pragmatique[5]. D'ailleurs, si les méthodes traditionnelles évoquées plus haut présentent des propriétés de convergence, il faut convenir que ces propriétés ne sont démontrables que pour l'optimisation de fonctions possédant les « bonnes » propriétés, et que ce sont rarement celles-là qui intéressent les ingénieurs[6].

Dès lors, le choix se situe entre des méthodes rigoureuses appliquées à des fonctions qui leur enlèvent ce caractère, et des méthodes inhabituelles par leur caractère intrinsèquement stochastiques, mais qui n'en possèdent pas moins quelques bonnes propriétés statistiquement démontrables (Goldberg, 1989).

---

[4] le site Internet de l'auteur (http://www.rennard.org) fournit des applet Java bien faites, en particulier pour ce problème du voyageur.
[5] c'est parfois plus difficile de convaincre un mathématicien.
[6] l'expérience montre que les fonctions à optimiser sont souvent définies par un mélange de paramètres continus et discrets ; elles sont rarement partout dérivables,  peuvent même présenter des discontinuités, et sont le plus souvent « multimodales », c'est-à-dire présentant de nombreux extremums relatifs (ne serait-ce qu'en raison de l'existence d'un bruit numérique que nous évoquerons plus loin) – ce qui leur enlève toute chance de convexité.

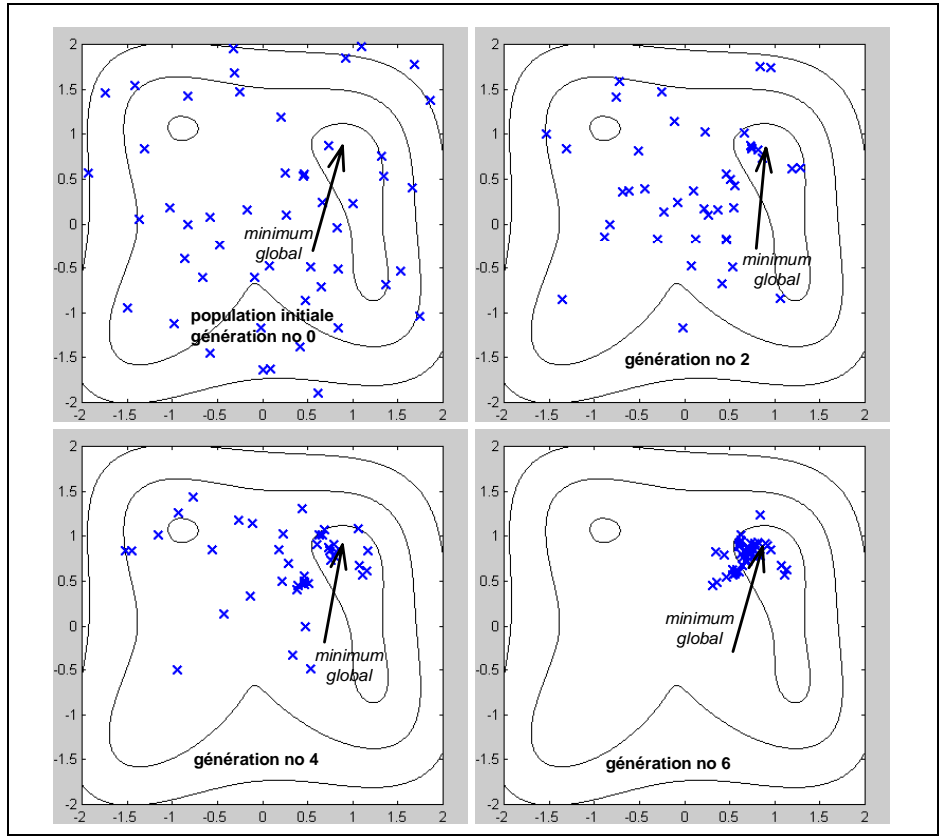

**Figure 5 :** *Exemple à 2 paramètres, montrant le déplacement de la population au fil des générations. Les lignes représentent les isovaleurs de la fonction coût, qui possède ici 3 minima locaux, plus un minimum global.*

### 1.4 Les opérateurs et l'algorithme génétiques de base.

Nous ne nous étendrons pas longuement sur la mécanique précise qui permet de construire chaque nouvelle génération à partir de la précédente : il existe de nombreuses excellentes références facilement accessibles sur ce sujet, dont il est par ailleurs difficile de faire une synthèse complète tant chacun fait preuve, en ce domaine, de créativité[7].

La méthode proposée à l'origine permet en tout cas de comprendre les principes de base des méthodes génétiques. Quels que soient le nombre des paramètres et leur type (discret ou continu), ils peuvent être codés en binaire et mis « bout à bout ». On obtient un codage binaire unique pour l'objet complet, codage appelé « chromosome ». La population complète contient donc N chromosomes, auxquels correspondent autant de variantes de l'objet qu'on est en train d'optimiser.

---

[7] un excellent tour d'horizon récent et en français des méthodes inspirées de la nature est (Siarry *et al.*, 2004). On y trouvera de très nombreuses références.

On définit deux « opérateurs »[8] qui permettront de faire « naître » de nouveaux individus (nouvelles variantes de notre objet) à partir de la population courante :

- le « croisement » génère deux nouveaux individus (des « enfants ») à partir de deux individus existants donnés (les « parents ») : on choisit pour cela aléatoirement un « site » (point sur le chromosome situé entre deux bits, après le premier et avant le dernier) ; chacun des chromosomes des parents est coupé à cet endroit  et les 4 morceaux sont croisés pour former les enfants, chacun héritant d'un morceau du chromosome de chacun de ses parents.

- la « mutation », qui génère un nouvel individu en modifiant de manière aléatoire l'un des bits du chromosome d'un individu existant.

En partant de la génération courante, on va générer la nouvelle génération en utilisant ces deux opérateurs :

- des couples sont formés pour procéder aux croisements. Le choix est aléatoire, mais construit de manière à favoriser les individus performants, qui auront donc statistiquement plus de descendants.

- des mutations sont introduites avec une probabilité faible (quelques mutations par génération).

- la population ainsi obtenue est plus nombreuse que la population d'origine. Après avoir évalué la performance des nouveaux individus générés (grâce à notre « boite noire » de calcul de coût), on ramène cette population à sa taille primitive par une opération de sélection, toujours menée de manière aléatoire, en éliminant de manière privilégiée, mais non systématique, les moins bons individus (au sens de la fonction coût choisie).

---

[8] à l'origine, Holland utilise un troisième opérateur, l'inversion, qui a peu été repris par la suite, en particulier pour le codage réel. Ce point mériterait d'être reconsidéré.

### 1.5 Pourquoi ça marche.

En quoi ce processus répond-il aux principes généraux définis plus haut (§2.3) ?

#### 1.5.1 Cas de deux parents performants et ressemblants.

Lorsqu'on croise de manière privilégiée les meilleurs individus, on a quelque chance qu'ils soient génétiquement voisins (proches dans l'espace des paramètres) ; leurs enfants seront « probablement » dans le même voisinage, et eux aussi de bonne qualité. Ils ont donc une faible probabilité d'être détruits par la sélection : cela contribue à augmenter la densité de population dans les zones favorables.

#### 1.5.2 Parents performants mais relativement différents.

Si les parents utilisés sont bons tous les deux, mais pas « voisins » dans l'espace des paramètres, leurs enfants seront probablement situés encore ailleurs dans cet espace de recherche. On espère que leurs qualités respectives pourront s'ajouter ou se multiplier : dans ce cas, on aura repéré une nouvelle zone intéressante. Dans le cas contraire, l'essai sera oublié grâce à la sélection.

On part donc de l'a priori suivant : par rapport à la recherche purement aléatoire, la probabilité de construire une très bonne solution est fortement augmentée si les candidats sont fabriquées par recombinaison d'éléments constitutifs de solutions qui sont elles-mêmes de bonne qualité. C'est en réalité une règle de bon sens, que chacun d'entre nous acquiert par l'expérience de la vie courante (certains utilisent le terme savant d' « heuristique »[9]). C'est aussi et nous y reviendrons, une règle de base des développements scientifiques, techniques et industriels des derniers siècles.

Cette heuristique n'est cependant pas sans fondements statistiques. L'expérience montre que les méthodes génétiques sont particulièrement efficaces sur les problèmes réels lorsque le nombre de paramètres est élevé (plusieurs dizaines par exemple). Dans une telle situation, s'il y a évidemment des paramètres qui présentent des effets antagonistes, ce n'est pas en pratique le cas majoritaire. Les croisements réalisés préférentiellement entre individus performants vont permettre de faire rapidement le tri et de repérer les zones que nous avons appelées plus haut « prometteuses », celles pour lesquels les qualités séparées s'ajoutent.

---

[9] d'après (Carré *et al.*, 1991) « une heuristique est une règle qu'on a intérêt à utiliser en général, parce qu'on sait qu'elle conduit souvent à la solution, bien qu'on n'ait aucune certitude sur sa validité dans tous les cas ».
Pour certains problèmes, on a le choix entre méthodes exactes et heuristiques. C'est le cas du problème du Master Mind : une stratégie « exacte » existe, qui permet de finir chaque partie en ne dépassant pas un certain nombre théorique de coups, lié au nombre de pions et de couleurs utilisés. Pour le même problème, il existe de bonnes heuristiques qui font de temps en temps un ou deux coups en trop. Sur plusieurs parties, elles dominent pourtant, statistiquement, la méthode dite exacte ! La différence est réelle, mais subtile : quel est *vraiment* l'objectif à atteindre, c'est souvent la question à laquelle il est le plus difficile de répondre lorsqu'on cherche à optimiser un objet …

C'est le sens général de la théorie des schémas exposée primitivement par Holland (Holland, 1975)[10]. Cette qualité des méthodes génétiques leur est spécifique : à notre connaissance, elle n'est présente dans aucune méthode locale[11].

### 1.5.3 Un ou deux parents peu performants.

Tous les autres croisements, et les mutations, servent à continuer l'exploration relativement aléatoire de l'espace de recherche, et à conserver une certaine « richesse génétique » qui, par croisement avec de bons individus, peut conduire à des améliorations supplémentaires.

### 1.5.4 Capacité à s'affranchir d'un bruitage de la fonction coût (Figure 6).

Suivant le type de méthode mise en œuvre pour résoudre le problème direct de la Figure 1 (évaluation de la fonction coût), le résultat obtenu  peut être perturbé par un certain bruit. Le cas typique est celui d'une résolution par éléments finis, lorsque les modifications de paramètres conduisent à la nécessité de refaire le maillage ; ou lorsque l'obtention de la solution nécessite la résolution par itérations d'un système non linéaire : le résultat obtenu est souvent entaché d'un bruit non négligeable, créateur de multiples minima locaux artificiels de la fonction coût. Là où une méthode locale simulant un gradient se laissera facilement « tromper », l'effet statistique de la recherche au travers d'une population est fortement correcteur[12].

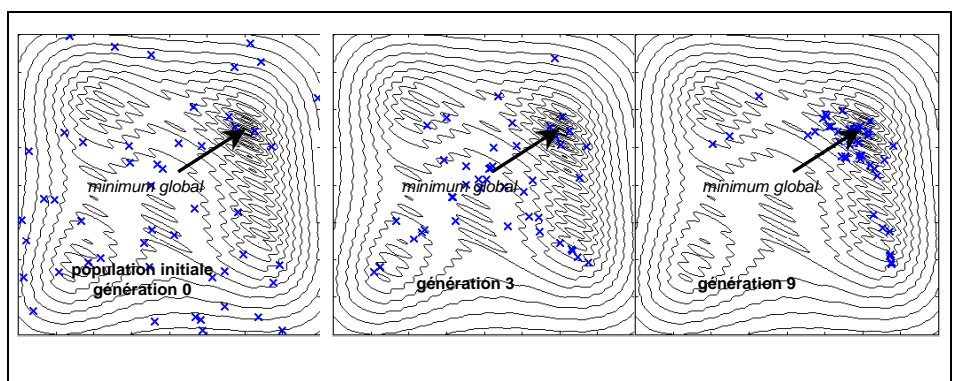

**Figure 6 :** *Convergence de la population vers l'optimum global malgré le « bruit » sinusoïdal ajouté (la fonction non bruitée est celle de la Figure 5).*

---

[10] théorie par ailleurs en partie contestée (Schoenauer, 2003).

[11] pour un nombre de paramètres significatif, c'est ce qui ferait qu'une méthode génétique à N individus serait plus efficace qu'une méthode déterministe locale lancée N fois à partir de chacun des N individus de la population génétique initiale.

[12] tant que la fréquence spatiale du bruit, dans l'espace des paramètres, reste grande devant celle de la fonction coût, et que le niveau de ce bruit reste raisonnable par rapport aux variations de la fonction coût, l'absence de corrélation entre ce bruit et la répartition de la population permet à cette population de *voir les variations utiles* de la fonction coût *malgré* la présence du bruit.

### 1.5.5 Exploration et exploitation.

Toutes les méthodes effectivement programmées ne fonctionnent pas exactement comme indiqué au §2.4, en particulier dans le cas de paramètres réels, mais le principe de base reste toujours le même : la méthode doit établir le juste équilibre entre la tendance à converger vers la meilleur solution déjà rencontrée (on parle d' « exploitation »), et la propriété à poursuivre la recherche dans d'autres zones de l'espace de recherche (on parle d' « exploration »). Autrement dit entre la simple amélioration par retouches légères de la meilleure solution connue, et la recherche de solutions nouvelles.

### 1.6 Evolution et finalité : quelle analogie avec la Nature ?

Beaucoup d'auteurs font une analogie directe entre l'évolution au sens de Darwin et ce type de méthode d'optimisation : une population d'une espèce donnée, placée dans un écosystème, évoluerait *pour* s'adapter à ce milieu. L'écosystème serait, dans notre problématique, représenté par la fonction coût, l'espèce par notre « population » en cours d'évolution artificielle.

Nous verrons plus bas que nous préférons pour notre part placer cette analogie avec les écosystèmes à un autre niveau ; une certaine prudence s'impose dans l'utilisation des métaphores, qui risquent d'induire des raisonnements erronés. Il n'y a en effet aucune *finalité* à l'évolution au sein d'un écosystème, qui ne poursuit aucun but ni explicite ni implicite ; l'espèce évolue effectivement d'une certaine manière *et* survit (elle n'évolue pas *pour* survivre, mais *parce que* – tous ses représentants n'étant pas identiques – certains ont survécu et d'autres pas[13]).

Les méthodes d'optimisation de type génétique seraient plutôt à mettre en parallèle avec les méthodes de sélection appliquées par les éleveurs ou les botanistes pour améliorer une race ou une espèce par rapport à un objectif spécifique qu'ils se fixent (obtenir un cheval puissant pour l'agriculture, un rapide pour la course, un docile et de petite taille pour la monte, …). Il est remarquable de constater que les éleveurs et les agriculteurs savaient travailler bien avant que les mécanismes de l'hérédité et de l'évolution soient expliqués. Cependant, l'accélération des évolutions génétiques induites par l'homme a été spectaculaire depuis Mendel et

---

[13] il est bien connu dans le grand public que les bactéries sont capables *de s'adapter pour* résister aux antibiotiques. La finalité de cette action « volontaire » apparaît explicitement dans le mode actif utilisé. Le caractère sournois des bactéries contre lesquelles l'industrie pharmaceutique doit lutter avec courage et constance est ainsi souligné. Cette vision animiste n'a (bien entendu ?) aucun fondement.

Darwin : la compréhension des mécanismes intimes a permis de perfectionner les « heuristiques »[14] !

## 2 Approche historique de l'évolution d'objets courants.

Nous venons de voir comment l'ingénieur peut transposer de manière efficace au monde technique des méthodes issues de la biologie. Essayons de renverser le point de vue, et de regarder le monde technique et son évolution depuis le 19ème siècle avec le regard d'un biologiste.

### 2.1 Exemple de l'automobile et de son écosystème.

Sur le modèle des arbres de l'évolution, par exemple celui des mammifères depuis l'ère jurassique, il est possible de représenter l'évolution d'objets techniques, bien entendu sur des périodes beaucoup plus courtes. L'automobile en est un exemple (Figure 7), qui pourrait être repris pour d'autres technologies, mais aussi pour les théories scientifiques, ou encore dans des domaines touchant aux méthodes ou à l'organisation sociale (agriculture, organisation du travail, …).

Poursuivons sur l'exemple de l'automobile. Pourquoi le modèle du haut de notre arbre, celui que nous voyons aujourd'hui dans les rues, n'a-t-il pas été proposé à la vente un siècle plus tôt, pour remplacer le modèle de 1898, très joli mais un peu ridicule avec ses grandes roues à rayons ?

La première réponse qui vient à l'esprit, c'est que la technologie nécessaire n'était pas disponible … C'est vrai mais ce n'est là qu'une partie de la réponse. Imaginons que nos bureaux d'étude d'aujourd'hui, avec tous leurs moyens modernes de calcul et disposant de toute l'expérience acquise depuis, essaient de concevoir un véhicule qui devrait être construit et utilisé en 1905. Ils auraient beaucoup de mal à faire mieux que ce qui avait été fait alors (du moins ce que nous en retenons aujourd'hui, c'est-à-dire ce qui avait fonctionné) ; autrement dit, il est probablement impossible de concevoir aujourd'hui des véhicules mieux adaptés à l'environnement de l'époque que ceux qui avaient alors été imaginés, construits, vendus, entretenus, … En tout cas notre Twingo d'aujourd'hui, proposée en 1905 n'aurait pas « survécu » plus de quelques heures et quelques kilomètres : avec les véhicules, ce sont aussi les clients qui ont évolué, le système social, les routes, les ateliers de construction et de réparation, les matériaux, les pneumatiques, les carburants, les lubrifiants …

_________________

[14] Pour Michel Serres, l'arrivée des OGM n'est ainsi qu'une étape, certes importante mais pas révolutionnaire, dans un processus de modification piloté par l'homme, et commencé il y a 15 millénaires.

La société dans son ensemble, agglomérat d'une multitude de systèmes complexes (techniques ou non, biologiques, sociaux, éventuellement intelligents), évolue du fait des évolutions de chacun de ses sous-systèmes, et réciproquement, chaque sous-système s'adaptant rapidement à ce qu'il perçoit de cette infinité d'évolutions particulières. Les êtres biologiques vivent dans un écosystème dont ils dépendent étroitement, qui les fait évoluer, et qu'ils font évoluer en retour. De la même manière, les objets techniques sont étroitement interdépendants et soumis à ce même processus de coévolution.

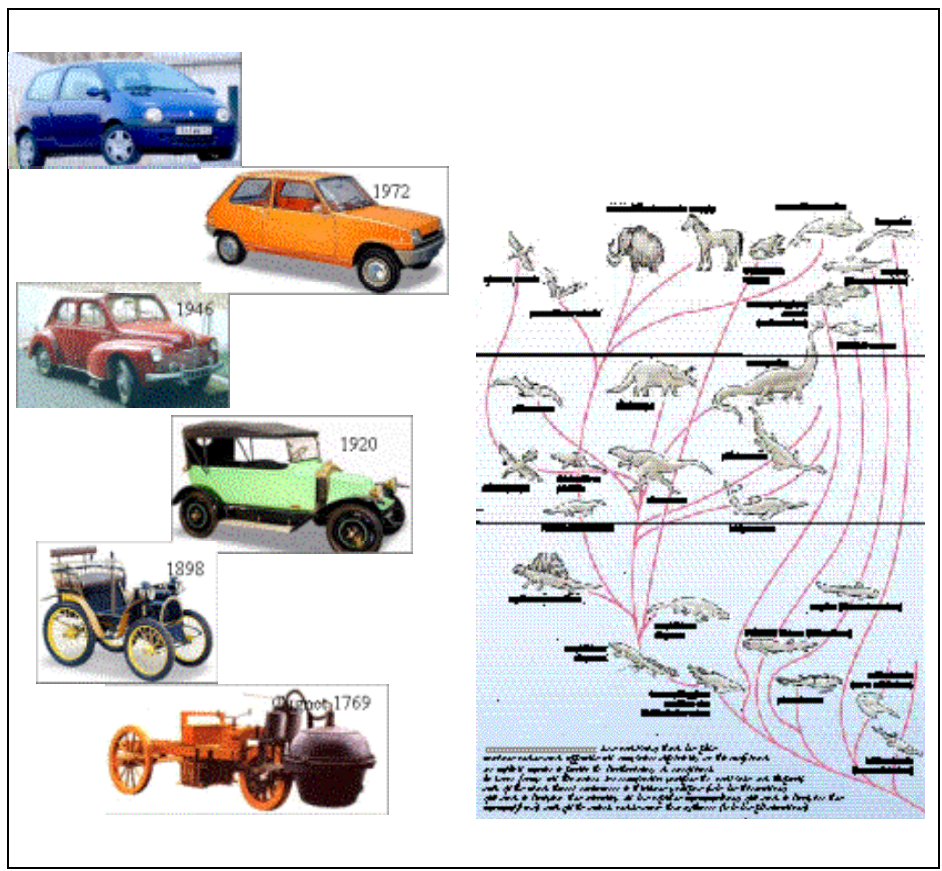

**Figure 7 :** *Deux arbres de l'évolution.*

## 2.2 Evoluer, mais vers quoi ?

### 2.2.1 Adaptation ou Progrès ?

Pouvons-nous vraiment dire que nous ne saurions pas faire mieux que les ingénieurs d'il y a 100 ans ? N'aurions-nous fait aucun « progrès » ? Nous sommes pourtant immergés dans un univers apparemment dominé par les progrès techniques. Dans beaucoup de cas, les observateurs que nous sommes confondent en réalité *progrès* et *adaptation* dynamique aux variations de l'écosystème technologique. Ces variations du contexte font qu'un objet parfaitement « adapté » cinq ans plus tôt l'est moins aujourd'hui ; le dernier modèle proposé, par contre, tient compte de ce nouveau contexte : s'il n'est pas objectivement « meilleur », il est objectivement mieux adapté. Cette différence d'adaptation est de ce fait ressentie comme positive (la fonction coût implicite est diminuée), et donc interprétée comme un progrès.

Comme observateurs, notre regard est donc piégé par une perspective trompeuse. C'est pire encore lorsque nous sommes nous-mêmes, en tant que scientifiques et ingénieurs, les moteurs de cette évolution : parce que nous « optimisons » avec effort et en utilisant toute notre intelligence, nous pensons aller vers un but volontairement fixé (et pour la raison évoquée ci-dessus, nous pensons que ce but est un progrès). Mais la direction de cette évolution n'est donnée qu'au niveau granulaire, celui du sous-sous-système. Cela ne saurait conduire à un quelconque pilotage de l'ensemble, celui de l'écosystème global (la société technique dans son ensemble), qui dérive en réalité de manière incontrôlée. L'extraordinaire coévolution qui permet à certains papillons d'adapter la forme de leur trompe à l'évolution de celle de la corolle de leurs fleurs préférées (ou aux dispositifs chasseurs de mines de lutter toujours plus efficacement contre des mines de plus en plus sophistiquées[15]) n'a pas empêché la fin des dinosaures.

### 2.2.2 Piloter l'évolution ?

En ce sens, le libéralisme pur, dans son refus des réglementations, est une aberration. Les réglementations[16] sont en réalité les seuls gouvernails explicites dont disposent les sociétés humaines pour tenter d'orienter leurs évolutions : le tri « naturel » par l'écosystème ne connaît que le temps présent, qui n'a que faire par exemple de considérations écologiques dont les enjeux sont sensibles sur le long terme[17]. Sans réglementations, le coût de la pollution n'interviendrait de manière

---

[15] la nature s'avère ici particulièrement performante, car l'insecte et la fleur ne sont pas, eux, conçus dans un même bureau d'étude. Quoique ? (voir les théories, populaires aux Etats-Unis, concernant l' « intelligent design ») (Wikipédia, Internet)

[16] au niveau technique, la normalisation. joue aussi ce rôle.

[17] certaines réglementations, tout aussi nécessaires et contestées par les mêmes courants d'idées, ne sont pas liées à cette idée d'anticipation, mais à des notions de progrès dans le

sensible sur l'évolution technique que beaucoup trop tard (sous forme catastrophique).

Prenons l'exemple des moyens de chauffage individuels dans les habitations neuves : malgré les chocs pétroliers successifs sur lesquels nous reviendrons, les logements neufs courants, et en particulier les maisons individuelles, restent très majoritairement chauffés à l'aide d'un vecteur qui est un scandale thermodynamique (le *convecteur* électrique). On a là typiquement un système quasi gratuit en investissement, mais dont le coût global (investissement plus fonctionnement sur 50 années) est énorme. Mais la majorité des clients concernés (jeune couple avec enfants) n'ont absolument pas les moyens d'investir dans un système plus performant dans la durée.

La seule méthode pour « corriger » l'écosystème et rendre massivement « vivables » d'autres moyens de chauffage, est d'agir au travers des réglementations : il faut d'une part un système de taxation écologique des « mauvais » dispositifs, et d'autre part une prise en charge par la collectivité du surcoût de l'investissement lié aux « bons » dispositifs, avec éventuellement un système de remboursement progressif lié à l'usage : le particulier concerné, comme la collectivité, s'y retrouveraient à long terme. De plus, l'écosystème réagirait spontanément dans le bon sens, puisque l'importance du marché ainsi ouvert à des dispositifs dont la diffusion était jusqu'alors confidentielle va conduire à la fois à une augmentation de leurs performances et à une baisse de leurs coûts.

### 2.3 Mécanismes de l'évolution technique.

Comment fonctionne cette coévolution technique, et jusqu'à quel point peut-on la comparer à l'évolution d'un écosystème biologique ?

On a vu au début de cet article qu'un bureau d'étude, confronté au défi d'améliorer une caractéristique d'un dispositif, va presque certainement parvenir à proposer une solution (même si lui-même ou ses concurrents avaient très bien conçu la génération présente de ce dispositif) : par exemple, tel nouveau matériau a, depuis, fait son apparition ; ou bien les performances du logiciel utilisé pour le calcul des structures est devenu plus précis et plus performant. Cela ne signifie pourtant pas de manière certaine que la solution proposée sera finalement un succès.

---

domaine éthique (abolition de l'esclavage, lutte contre l'exploitation des jeunes enfants et la domination des femmes par les hommes, droit du travail et droit syndical …).

### 2.3.1 Conception et innovation : la méthode des essais-erreurs en vraie grandeur.

En réalité, c'est en permanence et dans le monde entier que de très nombreux essais sont faits dans tous les domaines, pour mieux adapter les objets[18] aux contextes dans lesquels ils trouvent leur utilité ou pour en concevoir de nouveaux.

On peut mesurer cela aux statistiques d'entreprises nouvelles créées, sachant que chacune de ces créations repose bien souvent sur un produit jugé suffisamment prometteur pour que des personnes fassent le pari d'y consacrer beaucoup d'énergie et d'argent ; mais aussi aux statistiques des entreprises qui disparaissent, ce qui correspond souvent à un produit en « fin de vie » (remplacé par un produit nouveau mieux adapté, et fabriqué ailleurs), ou à une idée nouvelle qui n'a pas abouti.

Rien que pour la France, ce sont ainsi plusieurs dizaines de milliers d'entreprises qui naissent chaque année, et autant qui disparaissent (INSEE, Internet). Les entreprises bien établies connaissent aussi des produits qui réussissent, et d'autres (pourtant étudiés avec autant de soin) qui ne parviennent pas à s'imposer[19].

### 2.3.2 La métaphore génétique de l'évolution technique.

Ainsi, vu de manière « macroscopique », le processus qui conduit à l'évolution technique est tout à fait comparable à l'évolution d'une espèce biologique dans un écosystème : à partir d'une génération d'objets remplissant avec succès une fonction donnée (par exemple les automobiles de classe moyenne de l'année 1970), de multiples propositions nouvelles sont imaginées et éventuellement construites, dont une partie seulement sera retenue (non par décision explicite d'une autorité centrale, d'un « juge », mais de manière implicite par l'écosystème technique, dont nous chercherons plus bas à préciser la nature).

Pour proposer ces solutions nouvelles, on s'inspire surtout des meilleures réalisations connues (sélection avant reproduction), on cherche à cumuler les qualités (croisement : parfois, cela fonctionne, d'autres fois, de nouveaux défauts inattendus apparaissent …) ; de temps en temps, une idée tout à fait nouvelle est ajoutée, ou une « erreur » est commise (mutation). certains individus de cette nouvelle population sont mieux adaptés et sont conservés (par exemple, la « R5 » apparue en 1972), tandis que d'autres s'avèrent à l'usage moins adaptés et sont rapidement abandonnés (exemple de la R14).

---

[18] prendre ce mot dans un sens très large.

[19] nous utilisons ici le langage lié aux entreprises et aux produits, mais notre discours s'applique à d'autres domaines, par exemple au succès des idées scientifiques. Avant de publier, chaque chercheur essaie de nombreuses voies, qui lui semblent toutes aussi prometteuses au départ. Il n'en retient finalement que quelques-unes pour publication, mais peut-être qu'une seule est finalement validée par les pairs. Le processus de sélection se poursuit au-delà, car il est finalement rare qu'une publication ait des suites (reprise de nombreuses fois comme référence, invitation à des conférences, …) jusqu'à rester comme une étape importante et (re)connue dans l'*évolution* d'une discipline.

Remarquons que cette adaptation à l'écosystème n'est pas quantifiable a priori, mais seulement mesurable a posteriori, en constatant le succès (survie) ou la disparition des objets concernés. La « fonction coût » est donc implicite, exactement comme dans le cas de l'évolution biologique , et contrairement à ce qui se passe dans le processus d'évolution forcée par sélection explicite, utilisée par les éleveurs, les agriculteurs … et les méthodes génétiques d'optimisation.

### 2.3.3 L'écosystème dépend du lieu.

Comme dans la nature, l'adaptation d'un objet donné, et donc l'écosystème, est *fonction du lieu*. Pour rester sur l'exemple de l'automobile, les besoins ne sont évidemment pas les mêmes en ville, à la campagne ou à la montagne. Ils ne sont pas les mêmes en France et en Afrique centrale. Ce tout dernier exemple montre bien ce qu'est l'écosystème dans son sens technique : la différence d'adaptation des véhicules ne tient pas seulement à la longueur des trajets qu'il doit accomplir ou à leur profil routier, mais aussi au niveau de vie des acheteurs potentiels, à la qualité des routes et pistes, aux conditions climatiques, aux possibilités d'approvisionnement en pièces de rechange (peut-être simplement à l'existence ou non d'un réseau de la marque considérée), à la technicité des ateliers de réparation, à la composition plus ou moins régulière des carburants, aux réglementations en vigueur, etc.

Un exemple de coévolution qui se passe différemment suivant les lieux est typiquement celui du parc automobile d'un côté, des carburants de l'autre. Ces deux variétés d'objets techniques évoluent conjointement. La France est actuellement dominée par le Diesel et le « sans plomb » (ce qui n'était pas le cas 20 ans en arrière), en apparence pour des raisons fiscales et écologiques d'ailleurs contradictoires ; d'autres pays, de niveau de vie équivalent, donnent à ce jour un rôle beaucoup plus confidentiel au Diesel[20] ; au Brésil, la puissance de l'agriculture et la taille immense du pays permettent de cultiver de la canne à sucre destinée aux biocarburants de manière rentable (le prix de revient devient concurrentiel dès que le pétrole dépasse US$30 par baril) et l'essentiel du parc automobile est « Flex », c'est-à-dire bi-carburants (Chalmin) ; en Afrique centrale, les carburants sans plomb sont quasi inexistants, car totalement inadaptés au parc automobile existant dont la moyenne d'âge est de 20 années plus élevée qu'en Europe de l'Ouest (Africaclean, Internet) : les véhicules neufs ne peuvent donc être les mêmes, ce qui contribue à perpétuer la différenciation des deux types d'écosystèmes.

### 2.4 Les dynamiques de l'écosystème.

L'adaptation dépend donc aussi *du temps* puisque l'écosystème change. On peut illustrer les comportements dynamiques possibles par quelques schémas, avec la représentation que nous avions utilisée pour expliquer le processus d'optimisation ; cette représentation devient ici symbolique, puisque les « fonctions coût » liées au processus de coévolution sont implicites.

---

[20] ce qui montre que l'écosystème dépend aussi des réglementations, elles-mêmes liées au poids de « lobby », ou (parfois) à l'intérêt public à long terme.

### 2.4.1 Dynamique incrémentale continue *de l'écosystème (Figure 8).*

Le plus souvent, l'optimum pour une fonctionnalité donnée se déplace par incréments liés aux modifications apportées aux autres éléments de l'écosystème ; l'objet courant perd lentement son adaptation, tandis qu'il est petit à petit remplacé par des objets mieux adaptés mais assez peu différents[21]. On pourrait imaginer arriver globalement vers un quasi équilibre (au moyen âge, chaque génération vivait « presque » comme la précédente), ou encore à des cycles (comme pour les modes vestimentaires). Ce serait ignorer l'existence de ruptures.

### 2.4.2 Dynamique incrémentale à rupture *de l'écosystème (Figure 9).*

En effet, et même lorsque tous les objets d'un écosystème suivent une évolution incrémentale continue, on peut arriver à une situation de rupture : c'est le cas lorsque deux solutions coexistent et que l'une d'entre elles, au départ moins adaptée à une majorité de situations, finit par s'imposer. L'exemple typique, pour rester dans le domaine des transports, est celui du basculement entre traction à vapeur et traction électrique pour les chemins de fer. On n'a cessé durant toute la première moitié du vingtième siècle d'utiliser et d'améliorer les deux modes de traction ; à partir des années 60, la traction électrique a cependant fini par s'imposer[22].

La rupture, au lieu d'être ainsi progressive, peut aussi parfois être brutale. Ce peut être le cas lorsque – à force de déformations incrémentales – les minima locaux des fonctions d'adaptation, qui assurent la stabilité de l'écosystème, finissent par disparaître. Les désordres sociaux qui apparaissent de temps à autre correspondent à ces moments où les « réformettes » ne suffisent plus à adapter l'organisation sociale ; seul un grand bouleversement (ou plusieurs successifs) permet de retrouver un nouvel équilibre satisfaisant : les acteurs procèdent spontanément à une recherche d'autres solutions, qui peuvent être éloignées de la précédente (révolution).

---

[21] cependant, « à la longue », la transformation peut être considérable. Nous pouvons évoquer ici l'exemple du moulin à eau, inventé par les romains, mais massivement utilisé seulement à partir du 10ème siècle, et qui n'a ensuite jamais été abandonné comme source d'énergie ; suite à des évolutions sans vraie rupture, il est devenu aujourd'hui l'un des éléments des grands systèmes complexes que sont les centrales hydroélectriques.

[22] ce résultat n'était pas du tout évident au départ. A la fin du XIXème siècle, les partisans de la traction à vapeur étaient nombreux même pour …le métro de Paris (celui de Londres avait commencé ainsi, il est vrai 30 ans plus tôt), et la polémique était bien réelle. L'électrification du Paris-Lyon date de 1952, Paris-Le Havre 15 ans plus tard : la transition incrémentale à rupture s'est étalée sur plus d'un demi-siècle.

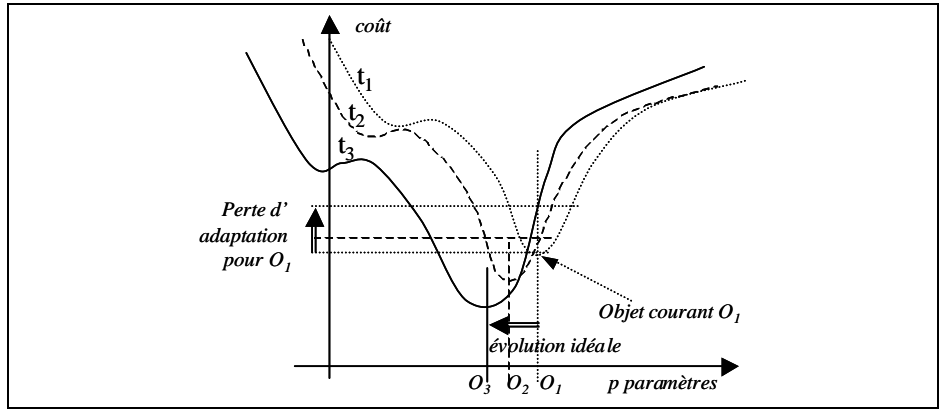

**Figure 8.** *Dynamique incrémentale continue*

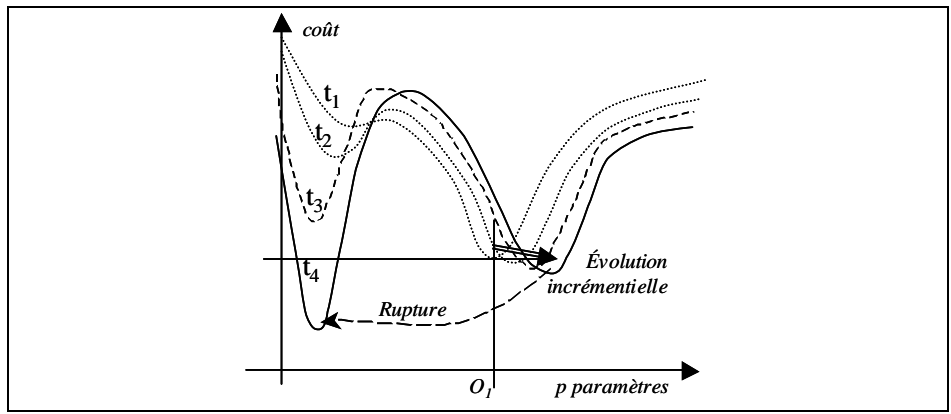

**Figure 9.** *Dynamique incrémentale à rupture.*

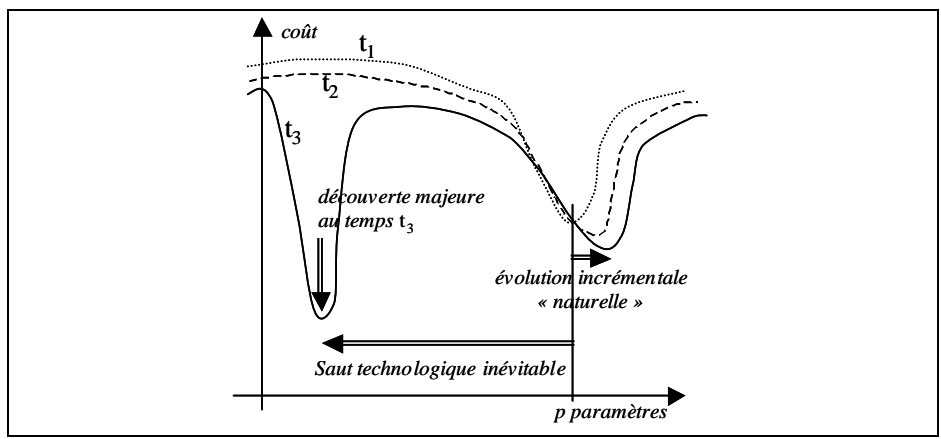

**Figure 10.** *Dynamique catastrophique.*

*2.4.3* Dynamique catastrophique *de l'écosystème (Figure 10).*

Mais il arrive aussi que l'écosystème soit soudain perturbé par un événement « extérieur » important. Dans le champ social, on peut évoquer une épidémie destructrice (peste noire au XIV[ème] siècle, SIDA aujourd'hui). Dans le domaine technique, il s'agit par exemple d'une découverte majeure (par exemple, le transistor), d'une catastrophe au sens propre (Tchernobyl) qui conduit à reconsidérer des choix fondamentaux, d'un changement de réglementation (exemple du chauffage cité plus haut), d'un événement géopolitique, …

La rupture et la catastrophe conduisent en réalité aux mêmes conséquences. Parfois, la perturbation reste « locale », et conduit seulement à la disparition d'une espèce au profit d'une autre. Mais il arrive aussi qu'il s'ensuive (par réaction en chaîne) des bouleversements rapides et brutaux pour une grande partie des constituants de l'écosystème, qui va peiner à retrouver une nouvelle situation quasi-stable du type *dynamique incrémentale continue*.

Ainsi de grands changements peuvent avoir une cause (on évoque souvent une météorite pour expliquer la disparition des grands dinosaures) ; mais une rupture du type précédent peut servir de déclencheur, et dans ce cas la « catastrophe » existe, mais il est illusoire d'en chercher la cause[23].

Le premier choc pétrolier, en 1973, a eu pour conséquence des modifications notables dans la hiérarchie des coûts (au sens économique du terme), conduisant à des bouleversements dans les structures des budgets des Etats, des entreprises et des particuliers, à la dévaluation immédiate de certains biens, à l'urgente nécessité de certains investissements (isolation thermique dans l'habitat), et à de nouvelles hypothèses de conception pour les transports, l'habitat, la production électrique, ... Du point de vue géopolitique global, il peut être considéré comme la conséquence d'une rupture ; vu de certains sous-systèmes (par exemple, les normes de construction), il apparaît comme une catastrophe externe.

Trente années après, il reste des traces – certes agaçantes – de ce bouleversement, et on peut lire ici ou là des critiques sévères du manque de prévoyance des architectes « qui construisaient n'importe comment ». Ce type de critiques est aisé après coup. Mais prenons l'exemple de l' « Ecole Centrale de Lyon », qui a justement été reconstruite en 1969. L'architecte qui aurait proposé un projet plus onéreux que ses collègues afin de réduire le coût à long terme du chauffage n'aurait évidemment pas été retenu pour ce marché ni pour aucun autre (l'écosystème l'aurait rejeté) : avec les hypothèses de l'époque, aucun investissement de ce type ne pouvait être justifié de manière raisonnée (ni financièrement, ni « écologiquement » car ce critère était alors très généralement ignoré), même à très long terme.

---

[23] ce n'est pas parce qu'un événement est brutal et spectaculaire qu'il a une cause. Ce que nous décrivons ici est bien entendu un autre regard sur les « bifurcations » qui existent dans les systèmes chaotiques, même si beaucoup d'auteurs s'obstinent à penser qu'un « déclencheur » extérieur reste indispensable (comme Pierre Grou dans Nottale, *Les arbres de l'évolution*). Ce n'est pourtant pas le cas dans une dynamique de rupture : une hypothèse vraisemblable serait ainsi qu'*il n'y a pas de cause* à la disparition des dinosaures, comme aussi à beaucoup de bouleversements historiques ou sociologiques majeurs.

Les cabinets d'architecte capables de concourir étaient ceux qui étaient adaptés à l'écosystème en place et qui par conséquent respectaient ses règles. Les individus qui contestaient les équilibres de l'époque ne jouaient aucun rôle économique ; dans l'écosystème, ils n'avaient alors conquis que quelques « niches » idéologiques.

### 2.4.4 Partage des ressources.

Un autre élément joue un rôle essentiel dans la dynamique de l'écosystème, qu'il s'agisse de biologie ou du monde technico-économique : une espèce n'est pas simplement présente ou absente, mais elle est présente avec un certain effectif. Cela conduit à des variations dans le temps assez complexes, puisque à la fois l'espèce évolue, mais aussi l'effectif de chacune des sous-espèces ainsi représentées.

Il est difficile de parler d'adaptation d'une espèce en tant que telle à un écosystème donné, sans tenir compte du nombre de ses représentants : l'adaptation étant définie au travers de la statistique de survie d'une génération à l'autre, elle sera différente suivant le nombre de représentants présents, avec le plus souvent des effets de seuils totalement non-linéaires, qui conduisent à des comportements imprévisibles. L'explosion de la population humaine depuis l'apparition de l'homme peut elle-même être analysée comme celle d'une espèce qui ne s'adapte plus par évolution génétique traditionnelle, mais par des moyens artificiels (maîtrise du feu et vêtements permettent de s'affranchir du climat, l'agriculture est liée à la multiplication de la densité de population, etc., jusqu'aux techniques les plus modernes). Elle accélère par là-même l'évolution de tous les écosystèmes avec lesquels elle est en contact, y compris bien sûr son environnement dans le sens traditionnel.

Parce que souvent plusieurs espèces voisines entrent en concurrence dans un écosystème donné (la petite voiture citadine a, en région parisienne, un nombre de clients annuels relativement stable : les différents modèles présents sur le marché vont se partager ces clients), on parle de « partage des ressources », ce qui permet d'estimer à un moment donné une limite haute pour le développement de telle ou telle espèce. Mais l'une des caractéristiques des systèmes complexes est d'évoluer vers davantage de complexité, vers une multiplication des espèces qui forment des réseaux imbriqués de ressources mutuelles : la notion de partage des ressources n'est en réalité valable qu'à court terme. En analysant, même finement, le marché des télécommunications en 1990, personne n'aurait pu imaginer ce qu'il est devenu 15 années plus tard, avec un téléphone par habitant de 8 ans ou plus, et une augmentation ahurissante de la fréquence de renouvellement de l'appareil lui-même : l'évolution du système a conduit à une multiplication soudaine des ressources.

### 2.5 Le mirage de la « fonction d'adaptation ».

Chacun sait dans les grandes lignes ce qu'est un champ électromagnétique, et nous pouvons donc le prendre comme métaphore. L'environnement d'un dispositif, et la manière avec laquelle cet environnement interagit sur la conception et les évolutions du dispositif, est comparable à l'action d'un champ électromagnétique sur une particule : le champ est une représentation pertinente de l'environnement de la particule étudiée ; mais en réalité, il représente l'action à distance de tout un ensemble d'autres particules, et il peut être amené à changer : indépendamment de la particule étudiée, ou à cause de sa présence.  De plus, suivant les caractéristiques propres de notre particule (chargée ou non, en mouvement ou non, …), ce ne sont pas les mêmes caractéristiques de ce champ qui devront être prises en compte.

Il en est de même de l'écosystème économique, technique, social et politique qui définit l'adaptation d'un dispositif donné : il s'agit de traduire l'interaction de notre dispositif (en tenant compte de ses caractéristiques particulières) avec un ensemble quasi infini d'autres dispositifs, techniques ou biologiques, éventuellement vivants et même intelligents. Cet environnement global est amené à varier, et à réagir dans le temps de manières différentes avec l'objet étudié, et ainsi à définir sa « survie », ou la manière avec laquelle il faudrait le faire évoluer pour assurer cette survie à ses successeurs.

La situation idéale pour le concepteur serait de connaître ce champ d'adaptation sous la forme d'une « fonction d'adaptation » explicite, modélisant l'ensemble de ces interactions en fonction des caractéristiques et paramètres définissant l'objet à concevoir, y compris le nombre d'exemplaires qu'on espère diffuser lorsqu'il s'agit d'un bien de consommation : le concepteur retrouverait alors le problème canonique d'optimisation (§2.1).

Malheureusement, il est totalement illusoire de poursuivre un tel objectif : les expériences de « vie artificielle » menées par simulation sur des systèmes dits « complexes » (mais obéissant en réalité à des règles de base extrêmement simples), démontrent qu'on ne parvient généralement pas à établir des lois de comportement, qui seules permettraient d'extraire pour l'un des sous-systèmes sa fonction d'adaptation (Holland, 1995[24] ; Santa Fe Institut, Internet). Alors même que les règles du jeu sont connues, simples et peu nombreuses, on en est réduit à rechercher de telles lois de comportement en appliquant des méthodes finalement plus proches de la sociologie que de la déduction mathématique.

---

[24] les références sur la simulation d'évolutions de systèmes complexes sont aussi nombreuses que celles concernant les méthodes génétiques ou évolutionnaires d'optimisation, et la plupart sont beaucoup plus récentes que 1995; l'intérêt de citer une nouvelle fois Holland est de montrer la continuité d'une pensée, et aussi de renvoyer indirectement aux travaux extrêmement variés réalisés à l'Institut Santa Fe.

## 3 Quelle démarche pour l'avenir ?

Les méthodes génétiques ou évolutionnaires utilisées pour l'optimisation sont généralement présentées comme une transposition des mécanismes biologiques qui ont permis l'évolution des espèces. Nous espérons avoir montré qu'il s'agit autant de la transposition et de la mise en forme systématique de mécanismes qui, implicitement, gouvernent les évolutions techniques, économiques et sociales (De Rosnay, 2005).

Notre conviction est finalement que seule une approche multidisciplinaire suffisamment large permettrait de mieux comprendre ce processus d'évolution général dont la biologie n'est qu'un cas particulier, et qui d'après les cosmologistes, a commencé dès le big-bang (Reeves, 2001 ; Nottale *et al*., 2000).

Parce que la biologie n'est qu'un cas particulier d'un processus plus général, il nous paraît difficile de continuer à utiliser le terme « évolution », trop nettement connoté, et nous proposons pour la suite de parler d'évolutique. Ce mot a une vie antérieure[25] mais semble être tombé en désuétude. Pour notre part, nous l'employons pour créer une association d'idées à partir des sonorités : d'un côté, évolution et darwinisme ; de l'autre monde scientifique et technique, mais aussi économique, sociologique et politique.

En quoi ce point de vue peut-il nous servir, puisque nous avons aussi dû admettre ne pas avoir accès à la « fonction d'adaptation » des objets que nous imaginons ?

### 3.1 Retour sur la démarche d'optimisation.

Nous proposons dans un premier temps quelques pistes de réflexion reliées à la démarche aujourd'hui classique d'optimisation par méthodes génétiques : à la lumière de l'évolutique, comment interpréter les fonctions coût que nous utilisons pour optimiser ; quelle distinction entre conception optimale et optimisation ?

---

[25] il semble avoir été utilisé pour la première fois en 1981 par J.-C. Rufin (l'auteur de livre grand public comme « Rouge Brésil »), dans son livre « L'évolution fixe » (Rufin, 1981). L'évolutique est pour lui l'étude de l'évolution des systèmes de systèmes, d'un point de vue théorique (c'est-à-dire distancié par rapport au support réel de ces systèmes, biologiques par exemple) ; il pense en particulier que les évolutions possibles d'un système de systèmes se passent dans un « espace mendeleïevien », c'est-à-dire qu'elles obéissent toujours à une classification du type « tableau périodique » : les possibles seraient en nombre fini, et tous prévisibles. Cette thèse n'a pas été reprise par d'autres auteurs.

Comment réinterpréter l'approche multi-objectifs de Pareto, aujourd'hui largement utilisée ? Quels pourraient être les apports de l'optimisation de systèmes ? Peut-on intégrer la dynamique de l'évolution dans nos travaux ?

Nous serons ici assez brefs : l'objectif n'est pas de clore le sujet, mais bien de l'ouvrir en provoquant des réactions des praticiens de l'optimisation.

### 3.1.1 La fonction coût.

Au niveau universitaire (nous ne savons pas tout ce que font les industriels, surtout les plus performants), il nous faut constater que les travaux consacrés à l'optimisation[26] font le plus souvent la part belle aux méthodes mises en œuvre (variantes subtiles, hybridations de méthodes, tests compliqués …) et laissent de côté l'approfondissement de la réflexion sur la ou les fonctions coûts utilisées. C'est évidemment regrettable, car si les résultats peuvent effectivement dépendre un peu de la méthode d'optimisation choisie, ils sont fondamentalement liés au choix des fonctions optimisées.

Pour la conception optimale, nous avons vu que l'idéal serait de pouvoir utiliser directement la fonction d'adaptation à l'écosystème, mais que nous n'y avons pas accès. Dans la démarche habituelle d'optimisation, la traduction des interactions complexes et nombreuses entre l'objet étudié et son environnement par une « simple » fonction – *objectif*, ou *coût*, ou *performance* – est une simplification pragmatique par rapport à la réalité. Il s'agit en fait de considérer qu'il est possible d'améliorer l'objet en ne prenant en compte qu'une toute petite partie des interactions existantes ; ou encore d'admettre implicitement que les paramètres sur lesquels on joue n'affectent pas notablement celles des interactions qui ne sont pas intégrées dans la fonction objectif.

On fait donc implicitement deux types d'hypothèses :   1 - le choix des paramètres est adéquat, ils vont permettre les modifications intéressantes de l'objet étudié ; 2 - les variations de la fonction coût choisie vont pour l'essentiel être homologues à celles de la fonction d'adaptation implicite (sans être confondues, ces fonctions ont en gros leur minimum au même endroit). Il faut être bien conscient que, plus on s'approchera d'un objet « idéal » pour un écosystème donné, plus il sera difficile de tenir ces deux hypothèses. Autrement dit, le choix de certains paramètres est forcément réducteur par rapport à la solution idéale, et pousser trop loin l'optimisation contrainte par ce choix préalable finit par perdre son sens. De même, à mesure que le choix affiné du paramétrage permet d'approcher la solution idéale, il faut prendre en compte le côté réducteur du choix particulier et simplificateur fait pour la fonction coût.

Un autre élément joue un rôle important, surtout si on prévoit de prendre en compte, dès la conception, la dynamique de l'écosystème. Il s'agirait d'établir une distinction claire entre les caractéristiques intrinsèques et extrinsèques qui interviennent pendant l'optimisation, au travers des paramètres ou des contraintes. Certaines des caractéristiques ne dépendent en effet que de l'objet lui-même, et ne risquent pas d'être modifiées par la dynamique de l'écosystème. Le plus souvent, les dimensions géométriques, du moins leurs détails, sont de cet ordre. D'autres sont à l'évidence directement liées à l'écosystème, le coût des matières premières en est un exemple, mais on peut aussi évoquer les matériaux disponibles et leurs propriétés limites, susceptibles de progresser, tous les éléments liés aux réglementations, etc.

---

[26] y compris les nôtres, bien entendu.

### 3.1.2 L'approche multiobjectifs de Pareto[27].

Les vrais problèmes ne font pas intervenir un seul critère, d'où les méthodes multi-objectif et en particulier les analyses de type « Pareto » (Zitzler, 1999) ; d'où aussi, lorsqu'on se contente d'une analyse mono-objectif, la recherche de plusieurs minima (le tri étant fait a posteriori en fonction d'autres critères).

Mais que signifient « plusieurs critères », par rapport à notre modèle évolutionniste, pour lequel on n'a bien qu'une seule fonction d'adaptation ? Deux situations types sont possibles :

- les différents objectifs considérés concernent des registres relativement indépendants, et ne sont pour l'essentiel pas gouvernés par les mêmes degrés de liberté durant la conception. Dès lors, deux optimisations peu interdépendantes peuvent être conduites, et les résultats ne sont pas contradictoires : dans chacun des domaines, les variations de la fonction d'adaptation sont essentiellement corrélées à l'un des objectifs considérés.

- l'approche de Pareto concerne la situation dans laquelle deux objectifs (ou plus) se trouvent être contradictoires, on risque de détériorer le premier si on améliore le second et réciproquement

En réalité, si l'écosystème destiné à recevoir l'objet en cours de conception était parfaitement connu, cette dernière situation ne se produirait pas. Mais au travers de la conception, on cherche également à obtenir des dispositifs adaptables à une large gamme de situations, ce qui est aussi une certaine garantie contre les effets de la dynamique des environnements. On est donc prêt à admettre de ne pas avoir l'objet idéal par rapport un écosystème précis, de manière à ce qu'il soit acceptable pour plusieurs.

Ainsi, la pondération des objectifs conduit à la construction d'une bonne approximation de la fonction d'adaptation pour une situation précise, alors que la frontière de Pareto permet de définir une gamme de solutions dans laquelle on peut ensuite piocher au cas par cas : c'est bien au moment du passage à un écosystème précis que les « autres critères » évoqués entrent en jeu. Dans certains domaines, ces gammes sont d'ailleurs commercialisées en tant que telles, les ressources disponibles se partageant d'elles-mêmes entre les variantes possibles, avec des succès pour chacune qui change effectivement avec l'écosystème (un modèle de voiture, vendu avec plusieurs niveaux de confort, et indépendamment plusieurs niveaux de performances, évidemment avec des prix en rapport : ces trois critères définissent une surface de Pareto. Suivant la géographie, le niveau de vie, la culture locale, les différentes combinaisons d'options sont plus ou moins adaptées).

---

[27] il est intéressant de noter que Vilfredo Pareto, nommé professeur à l'Université de Lausanne en 1893, économiste et sociologue connu pour avoir approfondi la notion d'optimum en économie, était en fait *ingénieur* de formation, et avait exercé comme tel pendant de longues années.

### 3.1.3 Quelques remarques complémentaires.

Nous évoquons ici tantôt la notion d'optimisation, tantôt celle de conception optimale. Il n'est peut-être pas toujours évident d'établir une telle distinction. L'optimisation serait plutôt l'opération simplifiée (canonique) ; la conception tient compte plus largement du contexte, et on peut faire appel plusieurs fois à des optimisations au sens réduit pendant le processus de conception optimale. Le contexte est alors provisoirement et artificiellement « figé », et traduit de manière simplifiée au travers d'une (ou quelques) fonction(s) coût.

L'idée générale est la même lorsqu'on parle d' « optimisation système ». Dans ce cas, le dispositif complet a pu être préalablement divisé en blocs fonctionnels, chacun ayant des interactions bien identifiées et généralement peu nombreuses avec d'autres blocs ; pour optimiser le système, on optimise tour à tour chacun de ces blocs, en considérant provisoirement que tous les autres le sont déjà, une convergence d'ensemble étant espérée et le plus souvent obtenue après quelques itérations. On a en fait ici un petit écosystème artificiel en cours d'évolution, et une possibilité simple d'étudier des phénomènes de coévolution.

## 3.2  L'évolutique  : une approche multidisciplinaire.

Pour finir, nous allons essayer de montrer l'intérêt qu'il y aurait à lancer une série de travaux de recherche interdisciplinaires mais étroitement coordonnés autour de l'évolutique : histoire des sciences et techniques, épistémologie, systématique[28], théorie de l'information, intelligence et vie artificielle, etc., … et bien sûr biologie.

Nous parlons d'intérêt, mais dans quel sens ? Il ne s'agit en tout cas pas seulement d'un jeu intellectuel. Nous pouvons ici reprendre l'analogie avec la biologie :  il y a toujours eu deux types d'évolutions obéissant aux mêmes mécanismes profonds. D'une part l'évolution « naturelle », d'autre part l'évolution forcée liée à l'élevage et à l'agriculture. Nous avons déjà souligné que le second est devenu beaucoup plus performant depuis que ces mécanismes ont été compris. L'industrie agroalimentaire « pilote » aujourd'hui l'évolution biologique pour un grand nombre d'espèces …

---

[28] classification des espèces, parfois appelée taxinomie.

Il en est de même pour les évolutions hors du champ biologique. L'approche historique, assistée de systématique, doit permettre de mieux comprendre la marche « naturelle » de l'évolutique. Forts de cette compréhension des mécanismes profonds mis en œuvre, les scientifiques et ingénieurs pourront affiner les mécanismes de l'innovation ou de la conception optimale, mieux anticiper les dynamiques des écosystèmes dont ils sont partie prenante, mieux organiser et utiliser la quantité énorme d'information dont ils disposent sur les systèmes qu'ils manipulent : bref, travailler comme des éleveurs performants d'idées, concepts, machines et systèmes. Et pourquoi pas, un jour, piloter un tout petit peu l'évolutique, qui pour l'instant n'est pas davantage finalisée que l'évolution …

### 3.2.1 Histoire des sciences et des techniques.

Pour comprendre l'évolution de l'espèce humaine, l'analogie avec les lois générales connues par ailleurs ne suffit évidement pas : il faut faire appel à l'archéologie et à l'anthropologie. De la même manière, l'évolutique doit faire appel à l'histoire des sciences et des techniques.

L'approche devrait tenir compte des objectifs spécifiques poursuivis : il s'agirait, sur quelques exemples précis choisis dans des domaines variés (grandes idées scientifiques, méthodes de l'ingénieur, machines, …) d'analyser les évolutions sur quelques siècles, en en construisant les arbres, de repérer les croisements qui ont été fructueux et aussi ceux qui ne l'ont pas été (fausses pistes), les ruptures, les catastrophes ; de dessiner les contours de ce qui paraît constituer les éléments essentiels de l'écosystème associé, et d'en estimer les variations liées à différents contextes (évolutique différenciée), …

Le repérage des « branches mortes » est d'un intérêt tout particulier. Elles correspondent à des objets qui ont été adaptés à une certaine époque, et qui ont disparu, le plus souvent suite à un processus de dynamique incrémentale à rupture (nous citions l'exemple des locomotives à vapeur). Or, le changement progressif de la fonction d'adaptation qui a conduit à une telle rupture peut se reproduire en sens inverse. L'exemple type est celui du moulin à vent, longtemps abandonné au profit d'autres sources d'énergie, et qui réapparaît aujourd'hui en force sous forme d'éoliennes. Et Loïc Péron, le grand navigateur, à propos de la conception d'un bateau révolutionnaire qui s'inspire largement d'un modèle assez classique en 1987, ne propose-t-il pas de « regarder en arrière pour mieux voir devant » ?

Ainsi les « niches » abandonnées doivent-elles êtres périodiquement revisitées, et leur recensement systématique ne serait-il pas une tâche improductive[29].

---

[29] dans le domaine des méthodes de l'ingénieur, une rupture forte est liée à l'apparition dans les années 70 des éléments finis et de la généralisation du calcul numérique, au détriment d'autres méthodes, le plus souvent analytiques, qui avaient pourtant fait leurs preuves. Mais depuis, le calcul formel a lui aussi fait des progrès considérables, et des méthodes abandonnées au profit des éléments finis pourraient dans certains cas s'avérer à nouveau compétitives.

### 3.2.2 Systématique appliquée aux sciences et techniques.

En biologie, la classification des espèces a précédé le darwinisme. Mais une fois acquise l'idée d'évolution, la taxinomie (et l'étude des fossiles) s'est révélée extrêmement utile pour en vérifier la généralité. En sciences et techniques, il n'existe pas de classification systématique des idées, concepts et objet ; les quelques essais très partiels réalisés démontrent tout de même la fécondité de la démarche[30].

L'une des idées de base de l'évolutique, c'est que les nouvelles idées et nouveaux objets sont essentiellement construits par croisement d'idées ou objets préexistants ; ou plus précisément par réarrangement de sous-systèmes constituant les idées ou objets existants. Dans « Hidden Order », Holland (Holland, 1996) parle de *building blocks*. Ces blocs de base, dont il existe plusieurs niveaux de complexité, ne sont pas forcément spécifiques à une discipline.

Des spécialistes de la systématique pourraient relever le défi d'appliquer leurs compétences à l'évolutique, afin d'aider épistémologistes, scientifiques et ingénieurs à établir une classification des objets (au sens large) qu'ils manipulent depuis le début de l'ère industrielle ; et surtout, au delà des cloisonnements disciplinaires, à identifier les premières familles de *building blocks*.

### 3.2.3 Génome implicite : rôle de la théorie de l'information.

En biologie, on sait aujourd'hui comment l'information est codée. Mais quels sont les codages en sciences et technique ? Comment formaliser le génome implicite qui fait que nos idées et objets se développent et évoluent effectivement en espèces ? En s'appuyant sur le travail de nos historiens et taxinomistes, il reviendrait à des spécialistes de la théorie de l'information et du codage de proposer des méthodologies, qui pourraient aussi s'appuyer sur les travaux déjà existants dans le domaine de la capitalisation de la connaissance.

Ce travail devrait intégrer ce que nous avons souligné à propos de branches provisoirement mortes de nos arbres d'évolutique. Certaines plantes ou être vivants sont capables de s'adapter de manière rapide à un changement brutal de leur écosystème[31] en utilisant le « souvenir » de situations comparables qui serait stocké dans leurs gènes. Peut-on imaginer un système d'information qui permette cela pour les objets manufacturés, ou pour les idées techniques ou scientifiques[32] ?

L'homme est la première espèce à s'être affranchie de l'un de nos principes de départ de l'optimisation par méthode génétique, qui est que c'est uniquement l'information stockée dans la population existant à un instant donné qui sert de mémoire sur l'évolution passée. Le langage et la tradition orale ont permis dans une première phase à l'expérience de se transmettre au-delà d'une génération : l'information quitte la sphère biologique, et dépasse l'apprentissage entre ascendants et descendants directs. L'écrit a permis ensuite une part de transmission même lorsque la connaissance n'était plus utilisée pendant plusieurs générations successives.

---

[30] pour les actionneurs électriques, qui vont du micro-moteur de montre à l'alternateur de centrale électrique, voir par exemple (Jufer, 2004).
[31] c'est-à-dire sans utiliser le processus normal de l'évolution, qui, s'étalant sur plusieurs générations, est par définition lent.
[32] dans les objets « nouveaux », la part immatérielle (commande, logiciel, …) est de plus en plus importante ; cela ne fait que renforcer le rôle que la théorie de l'information est appelée à jouer.

Dès lors, la question est de savoir comment utiliser toute l'expérience stockée sur le passé, comment la structurer pour qu'elle soit disponible et activée au bon moment. Une fois identifiés les éléments de base, ou *building blocks*, et construits les arbres évolutiques, c'est là essentiellement une question d'organisation pertinente de l'information : comment construire et expliciter les génomes de l'évolutique ?

**Conclusion.**

Nous évoquions au début de cet article la difficulté psychologique ressentie par l'ingénieur, mais surtout par le scientifique ou mathématicien, lorsqu'il doit admettre que, parfois, la recherche (partiellement) aléatoire de solutions est plus efficace que ce que permet le raisonnement. On peut aller plus loin encore dans la provocation, en regardant avec quelque recul les progrès scientifiques même les plus pointus, attachés aux grands noms de la physique. Un chroniqueur scientifique écrivait ainsi récemment dans un hebdomadaire grand public (Allègre, 2005) :

« La science est une aventure humaine collective, continue, avec des milliers de chercheurs qui contribuent chaque jour à des progrès. Des découvertes, il y en a des milliers chaque année dans tous les laboratoires du monde (…). De temps en temps, bien sûr, il y a des sauts un peu plus grands que les autres, mais ce sont souvent des germes qui suivent une cristallisation «en l'air». (…). Si Einstein n'avait pas *découvert* la relativité, un autre l'aurait fait. ».

Ce travail de fourmis qui consiste à essayer toutes sortes de solutions nouvelles, le plus souvent (pour ne pas dire toujours) par combinaison d'idées déjà existantes ressemble beaucoup à notre représentation canonique du problème d'optimisation, traité par une méthode génétique. Cela ne donne bien sûr pas chaque fois un résultat intéressant, mais les bons résultats ne sont accessibles qu'à ce prix. La publication des résultats de recherche devrait de ce fait toujours présenter les tentatives infructueuses qui ont finalement conduit au succès, après bien des tâtonnements ; elles nous en apprendraient plus sur le processus de découverte que la soi-disant construction directe par raisonnement pur qui est habituellement publiée !

Mais pour revenir à la relativité, il y a tout de même une preuve du génie[33] d'Einstein, c'est qu'il ait réussi, il y a juste un siècle, dans *plusieurs* domaines de la physique ; et qu'il ait *récidivé* des années plus tard !

---

[33] Dans la perspective évolutique, nous avons beaucoup de mal à situer les génies scientifiques : il semble plutôt que la connaissance progresse dès lors que la pression de recherche est suffisante. Y a-t-il là un paradoxe ? Nous laisserons cette question ouverte !